# Small Sample Inference for Generalization Error in Classification Using the CUD Bound

Eric B. Laber and Susan A. Murphy
Department of Statistics
University of Michigan
Ann Arbor, MI 48104
{ laber, samurphy }@umich.edu

## Abstract

Confidence measures for the generalization error are crucial when small training samples are used to construct classifiers. A common approach is to estimate the generalization error by resampling and then assume the resampled estimator follows a known distribution to form a confidence set [Kohavi 1995, Martin 1996, Yang 2006]. Alternatively, one might bootstrap the resampled estimator of the generalization error to form a confidence set. Unfortunately, these methods do not reliably provide sets of the desired confidence. The poor performance appears to be due to the lack of smoothness of the generalization error as a function of the learned classifier. This results in a non-normal distribution of the estimated generalization error. We construct a confidence set for the generalization error by use of a smooth upper bound on the deviation between the resampled estimate and generalization error. The confidence set is formed by bootstrapping this upper bound. In cases in which the approximation class for the classifier can be represented as a parametric additive model, we provide a computationally efficient algorithm. This method exhibits superior performance across a series of test and simulated data sets.

## 1 Introduction

Measures of uncertainty are particularly critical in the small sample setting where training set to training set variation is likely to be high. One example occurs in medical diagnostics in mental health in which one may want to classify patients as "low risk" or "high risk" for relapse based on observed medical history. In these settings the signal to noise ratio is likely to be small. Thus, both the estimated generalization error and the confidence in this estimate must be considered as the confidence might be very low due to the noise in the data. In addition, in this setting data must be collected from clinical trials and is subsequently expensive to obtain. The cost of these trials, coupled with notoriously low adherence rates, leads to data sets which are too small to admit a test set. Any inference for the generalization error must be derived from a single training set.

In this paper, we introduce a new method for obtaining confidence bounds on the generalization error in the small sample setting. The paper is organized as follows. Section 2 of the paper provides the framework for this research, and describes why this problem is difficult. We also survey several standard approaches to this problem and motivate the need for a new method. In section 3 we derive a new bound called the constrained uniform deviation bound (CUD bound hereafter) on the deviation between the estimated and true generalization errors and describe how to use this bound to form confidence sets. Section 4 gives an efficient algorithm for computing the new upper bound when the classifier can be represented as a parametric additive model. Section 5 describes an empirical study of several methods for constructing confidence sets and analyzes the results.

## 2 Framework

Our objective is to derive a confidence set for the generalization error under minimal assumptions. To this end, we only assume a training set $\mathcal{D} = \{(x_i, y_i)\}_{i=1,\ldots,n}$ of $n$ iid draws from (unknown) distribution $\mathcal{F}$ over $\mathcal{X} \times \mathcal{Y}$, where $\mathcal{X}$ is any feature space and $\mathcal{Y} = \{-1, 1\}$ is the label space. We also assume that there is a space of potential classifiers $\mathcal{G}$ from which we choose our classifier $f$ using some criterion that depends on our training set $\mathcal{D}$. Equally important is what we *do not* assume. We make no assumptions about the structure of $\mathcal{G}$, nor do we assume that the Bayes classifier belongs to $\mathcal{G}$. Finally, we do not assume the existence of an underlying "true" deterministic function $y = g(x)$ which generates the data.

Given a training set $\mathcal{D}$ we choose a classifier $f \in \mathcal{G}$ using some criterion (discussed in detail below). The classifier $f$ aims to predict the value of a label $y$ given feature $x$. Thus, a natural measure of the performance of $f$ is its *generalization error* $\xi(f) = \mathbb{E}_{(X,Y) \sim \mathcal{F}} \mathcal{I}\{f(X) \neq Y\}$. That is, the probability that $f$ will fail to correctly predict label $Y$ from feature $X$.

When data are abundant, inference for the generalization error is straightforward. In this setting, one partitions the available data into two sets $\mathcal{L}$ and $\mathcal{T}$ called a learning set and a testing set respectively. Estimation is done by choosing classifier $f \in \mathcal{G}$ using learning set $\mathcal{L}$ and then using empirical estimate $\hat{\xi}_{test}(f) = \frac{1}{\#\mathcal{T}} \sum_{(x_i, y_i) \in \mathcal{F}} \mathcal{I}\{f(x_i) \neq y_i\}$. The exact distribution of $\hat{\xi}_{test}$ is known and confidence sets can be computed numerically [Langford 2005(1)]. Moreover, $\hat{\xi}_{test}(f)$ is, under mild regularity conditions, asymptotically normal, making standard asymptotic inference applicable.

The convenient properties of $\hat{\xi}_{test}(f)$ are a direct consequence of the independent test set $\mathcal{T}$. We define the small sample setting as one in which the data are too meager to admit a test set. Without a test set one is forced to use the same data both to train and evaluate the classifier. Estimates and confidence sets for the generalization error must be constructed using resampling (e.g. bootstrap, cross-validation, etc.). However, when the training sets are small, the training set to training set distribution of these estimates may be highly non-normal, thus complicating the construction of confidence sets. This is particularly the case here as the generalization error $\xi(f)$ is a non-smooth function of $f$.

It is worth pointing out that there exist alternative motivations for seeking confidence sets in the absence of a testing set. The formation of a test set essentially "wastes" observations which are not used in the training of the classifier. The omission of these observations can significantly deteriorate the performance of the learned classifier. Another motivation is the existence of learning algorithms which assume that training accuracy correctly reflects generalization error when selecting a classifier. Using high quality bounds on the generalization error in the algorithm can boost performance [Langford 2005(2)] particularly when the training data is small.

**Relevant Work**

Much of the work on small sample inference for the generalization error has focused on point estimates via resampling. Popular point estimates include the .632+ estimator given by Efron and Tibshirani [Efron 1995], the $LV1O^*$ estimate given by Weiss [1991], and more recently the bolstered error estimate of Braga-Neto and Dougherty [Dougherty 2004]. A nice survey of generalization error estimates is given by Schiavo and Hand [Schiavo 2000].

In the literature, a variety of general approaches have been suggested for constructing confidence sets for the generalization error in the absence of a test set. We discuss them in turn.

1. *Known distribution.* One approach to this problem is to assume the resampled estimator follows a known distribution $\Phi$ [Kohavi 1995, Yang 2006]. Typically $\Phi$ is taken to be normal, so that a $1 - \delta$ confidence set for estimated generalization error $\hat{\xi}(f)$ of classifier $f$ is given by

$$\hat{\xi}(f) \quad \pm \quad \Phi^{-1}\left(\frac{\delta}{2}\right) \sqrt{\nu}/\sqrt{n^*},$$

where $\nu$ is some estimate of the standard error and $n^*$ is the effective sample size[1] used to construct the resampled estimator. This approach mimics the case where $\hat{\xi}(f)$ is constructed from a training set and then a normal approximation would have been justified for the estimator of the generalization error based on an independent test set.

2. *Bayesian Approach.* This approach mimics the case where $\hat{\xi}(f)$ is constructed from a training set and then using an independent test set, a Bayesian approach is used to construct a posterior confidence set. Here we specify a Beta prior $\pi$ on the generalization error $\xi(f)$ [Martin 1995].

$$\xi(f) \quad \sim \quad \beta(\gamma, \lambda).$$

If an independent test set $\mathcal{T} = \{(x_i, y_i)\}_{i=1,\ldots,m}$ were available then random variables $z_i = \mathcal{I}(f(x_i) = y_i)$ would be iid $bern(\xi)$. The posterior $P(\xi(f)|z_1, \ldots, z_m)$ is also a Beta distribution,

$$\xi(f)|z_1, \ldots, z_m \quad \sim \quad \beta(\gamma + m\hat{\xi}(f), \lambda + m - m\hat{\xi}(f))$$

Note that the posterior depends only on the number of misclassified points, $m\hat{\xi}(f)$ on the test set.

---
[1] Here we use effective sample size to mean the expected number of unique points used in estimation during a single step of a resampling procedure.

Suppose we do not have a test set. Then given resampled estimate $\hat{\xi}$, we can view $(n - n^*)\hat{\xi}(f)$ as the misclassified examples where $n$ and $n^*$ are the full and effective sample sizes. One can think of this as training on a set of size $n^*$ and testing on a set of size $n - n^*$. Using this idea, the resampled model is given by,

$$\begin{aligned}
\xi(f) &\sim \beta(\gamma, \lambda) \\
\xi(f)|z_1, \ldots, z_m &\sim \beta(\nu, \eta) \\
\nu &= \gamma + \hat{\xi}(f)(n - n^*) \\
\eta &= \lambda + (1 - \hat{\xi}(f))(n - n^*).
\end{aligned}$$

The posterior given above can be used to construct confidence intervals for $\xi(f)$.

3. *Direct Estimation.* In this approach we repeatedly resample the data and from each resampled dataset we construct a point estimate of the generalization error (this may require further resampling). This process generates a sequence of estimates for the generalization error. A confidence set can be constructed, for example, by taking the quantiles of the generated sequence of resampled estimators.

The methods described above can work well in some settings. Distribution-based methods tend to work well when distributional assumptions are met, providing tight confidence sets and accurate coverage. However, these methods can be non-robust to large deviations from distributional assumptions and can suffer from unacceptable type I errors. Direct estimation avoids distributional assumptions but is unstable for small samples because of non-smooth 0-1 loss. These problems are illustrated in the experiments sections of the paper.

Ideally confidence sets for the generalization error should satisfy the following criteria.

1. The confidence set should deliver the promised coverage for a wide variety of examples.

2. The confidence set should be efficiently computable.

3. The confidence set should be theoretically justified.

Given multiple algorithms for constructing confidence sets satisfying the above, algorithms that produce confidence sets of smallest size are preferred. We propose a new approach and compare this approach to the above methods using the first and second criteria in the experiments section.

## 3 The CUD Bound

We construct an upper bound on the deviation between the training error and generalization error of a classifier using ideas from the large literature on constructing bounds on the so-called *excess-risk* of a selected classifier [Shawe-Taylor 1998, Barlett and Mendelson 2006]. Also we allow the use of a convex surrogate as in Zhang [2004] for the 0-1 loss. In particular we use the idea of constructing a bound for the supremum of the deviation between the training error and generalization error over a small subset of the approximation space $\mathcal{G}$. This small subset is determined by the convex surrogate loss, $\mathcal{L} : (\mathcal{D}, f) \to \mathbb{R}$ used to construct the classifier ($\hat{f} = \arg\min_{f \in \mathcal{G}} \mathcal{L}(\mathcal{D}, f)$). A variety of surrogate loss functions may be used, a common loss function is squared error loss, given by $\mathcal{L}(\mathcal{D}, f) = \frac{1}{n}\sum_{(x_i,y_i) \in \mathcal{D}}(y_i - f(x_i))^2$. Some other examples are the hinge loss, logistic loss, exponential loss and penalized quadratic loss.

The intuition for the upper bound is as follows. Suppose we knew (we don't) the limiting value of $\hat{f}$, say $f^*$, as the amount of training data becomes infinite. $f^*$ may not be the Bayes classifier. Also notice that $f^*$ *does not* depend on the training set. In addition, since we can think of $\hat{f}$ as being concentrated near $f^*$, this motivates the use of $f^*$ as the anchor for the small subset over which we take the supremum. In particular, we take the supremum over the set of classifiers belonging to a neighborhood of $f^*$. The size of this neighborhood should depend on how far $\hat{f}$ is from $f^*$ in terms of the difference in loss, $\mathcal{L}(\mathcal{D}, \hat{f}) - \mathcal{L}(\mathcal{D}, f^*)$.

To fix notation, let $\hat{\xi}_{\mathcal{S}}(f)$ be the empirical error of $f$ on data set $\mathcal{S}$, that is,

$$\hat{\xi}_{\mathcal{S}}(f) = \frac{1}{\#\mathcal{S}} \sum_{(x_i, y_i) \in \mathcal{S}} \mathcal{I}\{f(x_i) \neq y_i\}$$

so that $\hat{\xi}_{\mathcal{D}}(\hat{f})$ is the *training error*. We have the following result.

**CUD Bound.** *If $\alpha_n$ is any positive function of the size $n$ of the training set $\mathcal{D}$ and $g(x) = (x+1) \wedge 1$ then:*

$$\left|\hat{\xi}_{\mathcal{D}}(\hat{f}) - \xi(\hat{f})\right| \leq \sup_{f \in \mathcal{G}} \left|\hat{\xi}_{\mathcal{D}}(f) - \xi(f)\right| g\big(\alpha_n\big(\mathcal{L}(\mathcal{D}, f^*) - \mathcal{L}(\mathcal{D}, f)\big)\big)$$

*Proof.* First notice that $g(x) \equiv 1 \,\forall\, x \in \mathbb{R}_+$. Then, since $\hat{f} = \arg\min_{f \in \mathcal{G}} \mathcal{L}(\mathcal{D}, f)$ we must have

$$\mathcal{L}(\mathcal{D}, f^*) - \mathcal{L}(\mathcal{D}, \hat{f}) \geq 0$$

so that,

$$g(\alpha_n(\mathcal{L}(\mathcal{D}, f^*) - \mathcal{L}(\mathcal{D}, \hat{f}))) = 1.$$

The result follows from noticing,

$$\begin{aligned}|\hat{\xi}_\mathcal{D}(\hat{f}) - \xi(\hat{f})| &= |\hat{\xi}_\mathcal{D}(\hat{f}) - \xi(\hat{f})| \times \\ &\quad g(\alpha_n(\mathcal{L}(\mathcal{D}, f^*) - \mathcal{L}(\mathcal{D}, \hat{f}))) \\ &\leq \sup_{f \in \mathcal{G}} |\hat{\xi}_\mathcal{D}(f) - \xi(f)| \times \\ &\quad g(\alpha_n(\mathcal{L}(\mathcal{D}, f^*) - \mathcal{L}(\mathcal{D}, f))),\end{aligned}$$

with the last inequality following because $\hat{f} \in \mathcal{G}$. □

The intuition here is that we consider a small neighborhood around $f^*$ with $\alpha_n$ chosen so that the size of the neighborhood shrinks at the same rate $\mathcal{L}(\mathcal{D}, \hat{f}) \to \mathcal{L}(\mathcal{D}, f^*)$ as the size of $\mathcal{D}$ tends to infinity. It is the function $g$ which restricts the domain over which the supremum is taken. In particular,

$$\begin{aligned}&|\hat{\xi}_\mathcal{D}(\hat{f}) - \xi(\hat{f})| \\ &\leq \sup_{f \in \mathcal{G}} |\hat{\xi}_\mathcal{D}(f) - \xi(f)| g(\alpha_n(\mathcal{L}(\mathcal{D}, f^*) - \mathcal{L}(\mathcal{D}, f))) \\ &\leq \sup_{f \in \mathcal{G}} |\hat{\xi}_\mathcal{D}(f) - \xi(f)| \mathcal{I}(\mathcal{L}(\mathcal{D}, f^*) \geq \mathcal{L}(\mathcal{D}, f) - \alpha_n^{-1}) \\ &= \sup_{f \in \mathcal{W}} |\hat{\xi}_\mathcal{D}(f) - \xi(f)|,\end{aligned}$$

where,

$$\mathcal{W} = \{f \in \mathcal{G} : \mathcal{L}(\mathcal{D}, f^*) \geq \mathcal{L}(\mathcal{D}, f) - \alpha_n^{-1}\}.$$

From the preceding set of inequalities we see that $g$ serves as a continuous approximation to the indicator function. The rate, $\alpha_n$, will depend on the complexity of $\mathcal{G}$ and the training set size, $n$. For example if $\mathcal{G}$ has low complexity such as a smoothly parameterized class with a finite dimensional parameter then the rate can be expected to be on the order of $n$. We now present the main result which follows immediately from the upper bound.

**Proposition 1.** *Let $\delta \in (0, 1]$, then if $\mathcal{Q}_{1-\delta}$ is the $1-\delta$ quantile of,*

$$\sup_{f \in \mathcal{G}} |\hat{\xi}_\mathcal{D}(f) - \xi(f)| g(\alpha_n(\mathcal{L}(\mathcal{D}, f^*) - \mathcal{L}(\mathcal{D}, f))),$$

*then $P(|\hat{\xi}_\mathcal{D}(\hat{f}) - \xi(\hat{f})| \leq \mathcal{Q}_{1-\delta}) \geq 1 - \delta$.*

Thus, the problem is reduced to computing $\mathcal{Q}_{1-\delta}$. This is done using the bootstrap.

### 3.1 Computing $\mathcal{Q}_{1-\delta}$

For any fixed $\delta$ we estimate $\mathcal{Q}_{1-\delta}$ with its bootstrap estimate $\hat{\mathcal{Q}}_{1-\delta}$. A bootstrap sample $\mathcal{D}^{(b)}$ from $\mathcal{D}$ is an iid draw of size $n$ from the distribution which puts equal mass on each point in $\mathcal{D}$. The bootstrap estimate is constructed by treating the original sample $\mathcal{D}$ as the true population and each bootstrap draw as a random draw of size $n$ from the true population. The mapping between population and bootstrap quantities is straightforward and given in the table below. The reader is referred to [Efron 1994] or [Hall 1992] for details.

| Population | Bootstrap |
|---|---|
| $\xi(f)$ | $\hat{\xi}_\mathcal{D}(f)$ |
| $f^*$ | $\hat{f}$ |
| $\mathcal{L}(\mathcal{D}, )$ | $\mathcal{L}(\mathcal{D}^{(b)}, )$ |
| $\hat{\xi}_\mathcal{D}(f)$ | $\hat{\xi}_{\mathcal{D}^{(b)}}(f)$ |

Using the above, the procedure to compute $\hat{\mathcal{Q}}_{1-\delta}$ is as follows.

1. Construct $B$ bootstrap datasets $\mathcal{D}^{(1)}, \ldots, \mathcal{D}^{(B)}$ from $\mathcal{D}$.

2. For $b = 1, \ldots, B$ compute:

$$\begin{aligned}\mathcal{Q}^{(b)} &= \sup_{f \in \mathcal{G}} |\hat{\xi}_{\mathcal{D}^{(b)}}(f) - \hat{\xi}_\mathcal{D}(f)| \times \\ &\quad g(\alpha_n(\mathcal{L}(\mathcal{D}^{(b)}, \hat{f}) - \mathcal{L}(\mathcal{D}^{(b)}, f))).\end{aligned}$$

3. Set $\hat{\mathcal{Q}}_{1-\delta}$ to the $1 - \delta$ sample quantile of $\mathcal{Q}^{(1)}, \ldots, \mathcal{Q}^{(B)}$.

We now substitute $\hat{\mathcal{Q}}_{1-\delta}$ into Proposition 1 in place of $\mathcal{Q}_{1-\delta}$ to construct the confidence set.

Initially the computation of the sup in step 2 above appears to be computationally intensive. Depending on the form of classifier space $\mathcal{G}$ and loss function $\mathcal{L}(, )$ taking this sup may prove to be computationally infeasible. However, when the loss function $\mathcal{L}(, )$ is convex and the classifier is approximated by a parametric additive model, computation is straightforward by way of a branch and bound algorithm which we discuss next.

## 4 An Efficient Algorithm For Parametric Additive Models

In this section we consider the space of classifiers

$$\mathcal{G} = \{f(x) = sign\Big(\sum_{i=1}^p \beta_i b(\gamma_i, x)\Big), \beta_i, \gamma_i \in \mathbb{R} \; \forall i\},$$

where $b(, )$ are basis functions indexed by the $\gamma_i$'s and the $\beta_i$'s are the basis coefficients. We also assume a convex surrogate loss function $\mathcal{L}(\mathcal{D}, f)$ where convexity refers to convexity in $\boldsymbol{\beta} = (\beta_1, \ldots, \beta_p)$ but not necessarily in $\boldsymbol{\gamma} = \gamma_1, \ldots, \gamma_p$. We assume that given training

set $\mathcal{D}$ we choose classifier

$$\hat{f}(x) = f(x; \hat{\boldsymbol{\beta}}, \hat{\boldsymbol{\gamma}}) = \arg\min_{\boldsymbol{\beta}, \boldsymbol{\gamma}} \mathcal{L}(\mathcal{D}, f(; \boldsymbol{\beta}, \boldsymbol{\gamma}))$$
$$= \arg\min_{\boldsymbol{\beta}} \left(\arg\min_{\boldsymbol{\gamma}} \mathcal{L}(\mathcal{D}, f(; \boldsymbol{\beta}, \boldsymbol{\gamma}))\right)$$
$$= \arg\min_{\boldsymbol{\beta}} \mathcal{L}(\mathcal{D}, f(; \boldsymbol{\beta}, \hat{\boldsymbol{\gamma}}))$$

We now hold $\hat{\boldsymbol{\gamma}}$ fixed and compute the upper bound as a supremum over $\boldsymbol{\beta}$.

**CUD Bound (Parametric Additive Models).** If $\alpha_n$ is any positive function of the size $n$ of the training set $\mathcal{D}$ and $g(x) = (x+1) \wedge 1$ then:

$$\left|\hat{\xi}_{\mathcal{D}}(f(; \hat{\boldsymbol{\beta}}, \hat{\boldsymbol{\gamma}})) - \xi(f(; \hat{\boldsymbol{\beta}}, \hat{\boldsymbol{\gamma}}))\right|$$

is bounded above by:

$$\sup_{\boldsymbol{\beta} \in \mathbb{R}^p} \left|\hat{\xi}_{\mathcal{D}}(f(; \boldsymbol{\beta}, \hat{\boldsymbol{\gamma}})) - \xi(f(; \boldsymbol{\beta}, \hat{\boldsymbol{\gamma}}))\right|$$
$$\times g(\alpha_n(\mathcal{L}(\mathcal{D}, f(; \boldsymbol{\beta}^*, \boldsymbol{\gamma}^*)) - \mathcal{L}(\mathcal{D}, f(; \boldsymbol{\beta}, \hat{\boldsymbol{\gamma}}))))$$

The supremum is difficult to calculate because of the term with the absolute values; it is the absolute difference between two sums of indicator functions and is hence both non-smooth and non-convex. We now discuss how to transform this problem into a series of convex optimization problems with linear constraints.

To begin, suppose $\mathcal{D}^{(b)}$ is a bootstrap dataset drawn from $\mathcal{D}$. For any training point $(x_i, y_i)$ in $\mathcal{D}$, let $\phi_i$ be the number of copies of $(x_i, y_i)$ in $\mathcal{D}^{(b)}$. Then $0 \leq \phi_i \leq n$ for all $i$ and the distribution of each $\phi_i$ is binomial with size $n$ and probability $\frac{1}{n}$. Then we can write,

$$n\left|\hat{\xi}_{\mathcal{D}^{(b)}}(f(; \boldsymbol{\beta}, \hat{\boldsymbol{\gamma}})) - \hat{\xi}_{\mathcal{D}}(f(; \boldsymbol{\beta}, \hat{\boldsymbol{\gamma}}))\right|$$
$$= \left|\sum_{(x_i, y_i) \in \mathcal{D}} (\phi_i - 1)\mathcal{I}\{f(x_i; \boldsymbol{\beta}, \hat{\boldsymbol{\gamma}}) \neq y_i\}\right|$$
$$= \left|\sum_{(x_i, y_i) \in \mathcal{D}^{(b)} - \mathcal{D}} (\phi_i - 1)\mathcal{I}\{f(x_i; \boldsymbol{\beta}, \hat{\boldsymbol{\gamma}}) \neq y_i\}\right|.$$

The above term does not depend on points with $\phi_i = 1$, that is, those points which were selected exactly once in the bootstrap resampling. We let $\mathcal{D}^{(b)} - \mathcal{D}$ denote the set of training points $(x_i, y_i)$ that satisfy $\phi_i \neq 1$.

Notice that the number of points in $\mathcal{D}^{(b)} - \mathcal{D}$ is necessarily smaller than $n$ which translates into computational savings. To see this, we will need to make use of the following equivalence relation.

Given $\boldsymbol{\beta}^1, \boldsymbol{\beta}^2 \in \mathbb{R}^p$ and $\mathcal{S}$ a subset of $\mathcal{D}$, we say $\boldsymbol{\beta}^1$ is congruent to $\boldsymbol{\beta}^2$ modulo $\mathcal{S}$ if $f(x; \boldsymbol{\beta}^1, \hat{\boldsymbol{\gamma}}) = f(x; \boldsymbol{\beta}^2, \hat{\boldsymbol{\gamma}})$ for all $x$ such that $(x, y) \in \mathcal{S}$ for some $y$ (i.e. $\boldsymbol{\beta}^1$ and $\boldsymbol{\beta}^2$ lead to the same classification on $\mathcal{S}$). Congruency modulo $\mathcal{S}$ defines an equivalency relation and hence partitions $\mathbb{R}^p$. For any fixed bootstrap dataset $\mathcal{D}^{(b)}$ let $\mathcal{S}$ be the collection of distinct points in $\mathcal{D}^{(b)} - \mathcal{D}$ and let $\mathcal{M}_1, \ldots, \mathcal{M}_R$ be the subsequent equivalence classes. Then, for any equivalence class $\mathcal{M}_i$,

$$\left|\hat{\xi}_{\mathcal{D}^{(b)}}(f(; \boldsymbol{\beta}, \hat{\boldsymbol{\gamma}})) - \hat{\xi}_{\mathcal{D}}(f(; \boldsymbol{\beta}, \hat{\boldsymbol{\gamma}}))\right| \equiv C(\mathcal{M}_i) \forall \boldsymbol{\beta} \in \mathcal{M}_i,$$

where $C(\mathcal{M}_i)$ is a constant. Then referring back to step 2 of the bootstrap procedure outlined in section 3 we can write,

$$\mathcal{Q}^{(b)} = sup_{\boldsymbol{\beta} \in \mathbb{R}^p} \left|\hat{\xi}_{\mathcal{D}}(f(; \boldsymbol{\beta}, \hat{\boldsymbol{\gamma}}) - \hat{\xi}_{\mathcal{D}^{(b)}}(f(; \boldsymbol{\beta}, \hat{\boldsymbol{\gamma}}))\right|$$
$$\times g(\alpha_n(\mathcal{L}(\mathcal{D}^{(b)}, f(; \hat{\boldsymbol{\beta}}, \hat{\boldsymbol{\gamma}})) - \mathcal{L}(\mathcal{D}^{(b)}, f(; \hat{\boldsymbol{\beta}}, \hat{\boldsymbol{\gamma}}))))$$
$$= sup_i C(\mathcal{M}_i)$$
$$\times \sup_{\boldsymbol{\beta} \in \mathcal{M}_i} g(\alpha_n(\mathcal{L}(\mathcal{D}^{(b)}, f(; \hat{\boldsymbol{\beta}}, \hat{\boldsymbol{\gamma}})) - \mathcal{L}(\mathcal{D}^{(b)}, f(; \boldsymbol{\beta}, \hat{\boldsymbol{\gamma}}))))$$
$$= \sup_i C(\mathcal{M}_i)$$
$$\times g(\alpha_n(\mathcal{L}(\mathcal{D}^{(b)}, f(; \hat{\boldsymbol{\beta}}, \hat{\boldsymbol{\gamma}})) - \inf_{\boldsymbol{\beta} \in \mathcal{M}_i} \mathcal{L}(\mathcal{D}^{(b)}, f(; \boldsymbol{\beta}, \hat{\boldsymbol{\gamma}})))),$$

with the last equality following from the monotonicity of $g$. Since $\mathcal{L}$ is convex in $\boldsymbol{\beta}$ this would be a series of convex optimization problems if membership in $\mathcal{M}_i$ could be expressed as a set of convex constraints. We now show how this can be done.

Let $\boldsymbol{\beta}_{\mathcal{M}_i}$ be a representative member of equivalence class $\mathcal{M}_i$. Then for any other $\boldsymbol{\beta} \in \mathcal{M}_i$ we must have,

$$f(x_j; \boldsymbol{\beta}_{\mathcal{M}_i}, \hat{\boldsymbol{\gamma}}) \sum_{k=1}^{p} \beta_k b(\hat{\gamma}_k, x_j) \geq 0$$

for all $x_j$ such that $(x_j, y) \in \mathcal{D}^{(b)} - \mathcal{D}$ for some $y$. Since $f(x_j; \boldsymbol{\beta}_{\mathcal{M}_i}, \hat{\boldsymbol{\gamma}})$ does not depend on $\boldsymbol{\beta}$ we see that membership to $\mathcal{M}_i$ is equivalent to satisfying a series of linear constraints, which is clearly convex. We have shown that calculation of $\mathcal{Q}^{(b)}$ amounts to a series of $R$ convex optimization problems.

In practice, even though $n$ is small (e.g. $n \leq 50$) $R$ can be very large (on the order of $2^n$ in the worst case) making direct computation of all $R$ convex optimization problems infeasible. Fortunately, exhaustive computation *can* be done using a branch and bound algorithm.

To use branch and bound we must recursively partition the search space. To do this we arbitrarily label the feature vectors in $\mathcal{D}^{(b)} - \mathcal{D}$ by $x_1, \ldots, x_d$. The root of the tree represents all of $\mathbb{R}^p$. The first left child represents the set of all $\boldsymbol{\beta} \in \mathbb{R}^p$ so that $\sum_{k=1}^{p} \beta_k b(\hat{\gamma}_k, x_1) \geq 0$ while the first right child represents all $\boldsymbol{\beta} \in \mathbb{R}^p$ so that $\sum_{k=0}^{p} \beta_i b(\hat{\gamma}_k, x_1) < 0$. In general, if $\mathcal{S}$ is a region defined by a node at level $j$ of the tree, then its left child represents the subspace of

$\mathcal{S}$ which satisfies $\boldsymbol{\beta} \in \mathcal{S}$ and $\sum_{k=0}^{p} \beta_k b(\hat{\gamma}_k, x_{j+1}) \geq 0$, similarly, its right child is the subspace of $\mathcal{S}$ which satisfies $\boldsymbol{\beta} \in \mathcal{S}$ and $\sum_{k=1}^{p} \beta_k b(\hat{\gamma}_k, x_{j+1}) < 0$. Notice that the terminal nodes of this tree are either infeasible or one of the equivalence classes $\mathcal{M}_1, \ldots, \mathcal{M}_R$.

To complete the branch and bound algorithm we need to define upper and lower bounds on the value of objective function,

$$\mathcal{O}(\boldsymbol{\beta}) = \left|\hat{\xi}_{\mathcal{D}^{(b)}}(f(;\boldsymbol{\beta},\hat{\boldsymbol{\gamma}})) - \hat{\xi}_{\mathcal{D}}(f(;\boldsymbol{\beta},\hat{\boldsymbol{\gamma}}))\right| \times$$
$$g(\alpha_n(\mathcal{L}(\mathcal{D}^{(b)}, f(;\hat{\boldsymbol{\beta}},\hat{\boldsymbol{\gamma}})) - \mathcal{L}(\mathcal{D}^{(b)}, f(;\boldsymbol{\beta},\hat{\boldsymbol{\gamma}})))).$$

In particular, given region $\mathcal{S}$ of $\mathbb{R}^p$ defined by a node on the tree representing our partition, we require an upper bound $\mathcal{U}(\mathcal{S})$ on $\mathcal{O}(\boldsymbol{\beta})$,

$$\sup_{\boldsymbol{\beta} \in \mathcal{S}} \mathcal{O}(\boldsymbol{\beta}) \leq \mathcal{U}(\mathcal{S}).$$

The upper bound $\mathcal{U}(\mathcal{S})$ is constructed by bounding the two terms in $\mathcal{O}(\boldsymbol{\beta})$ separately. An upper bound on $|\hat{\xi}_{\mathcal{D}^{(b)}}(f(;\boldsymbol{\beta},\hat{\boldsymbol{\gamma}})) - \hat{\xi}_{\mathcal{D}}(f(;\boldsymbol{\beta},\hat{\boldsymbol{\gamma}}))|$ can be obtained by noticing that if $\mathcal{S}$ is a region of $\mathbb{R}^p$ defined by a node at level $j$, then the classification of the first $j$ points in $\mathcal{D}^{(b)} - \mathcal{D}$ is fixed on $\mathcal{S}$. The upper bound is constructed by assuming the worst performance possible on the remaining $d-j$ points. To bound the second term, we compute,

$$sup_{\boldsymbol{\beta} \in \mathcal{S}} g(\alpha_n(\mathcal{L}(\mathcal{D}^{(b)}, f(;\hat{\boldsymbol{\beta}})) - \mathcal{L}(\mathcal{D}^{(b)}, f(;\boldsymbol{\beta},\hat{\boldsymbol{\gamma}})))),$$

which is a convex optimization problem. The upper bound $\mathcal{U}(\mathcal{S})$ is the product of these two upper bounds.

In addition, we require the following lower bound,

$$L(\mathcal{S}) \leq sup_{\boldsymbol{\beta} \in \mathcal{S}} \mathcal{O}(\boldsymbol{\beta}).$$

The lower bound $L(\mathcal{S})$ is obtained by plugging any feasible point in $\mathcal{S}$ into $\mathcal{O}(\boldsymbol{\beta})$. In practice, a natural choice is the *argsup* of the second term in the objective function, which has already been computed during the construction of $\mathcal{U}(\boldsymbol{\beta})$.

This algorithm running on a standard desktop with a sample size of $n = 50$ and using a total of 500 bootstrap samples to construct a confidence set runs in a only a few minutes. While this algorithm is still an NP-hard problem,[2] in practice it is significantly less computationally intensive than evaluating all possible classifications. The reason is that the function $g$ allows for significant reduction of the search space by restricting attention only to classifiers within a fixed distance of selected the classifier $\hat{f}$. To see this, notice that in the construction of $\mathcal{U}(\mathcal{S})$, if the term

$$sup_{\boldsymbol{\beta} \in \mathcal{S}} g(\alpha_n(\mathcal{L}(\mathcal{D}^{(b)}, f(;\hat{\boldsymbol{\beta}})) - \mathcal{L}(\mathcal{D}^{(b)}, f(;\boldsymbol{\beta},\hat{\boldsymbol{\gamma}})))),$$

is non-positive we can remove the region $\mathcal{S}$ from our search space since we know $\mathcal{O}(\hat{\beta}) \geq 0$.

## 5 Experiments

In this section we describe a set of experiments designed to test the coverage and diameter of confidence sets constructed using the CUD bound. To form a baseline for comparison we also construct confidence sets using a handful of methods found in the literature. These consist of two non-parametric bootstrap methods, a normal approximation using the bootstrap [Kohavi 1995], a normal approximation using CV [Yang 2006], a generalization of a Bayesian approach [Martin 1995], and the inverted binomial [Langford 2005].[3] An online supplemental appendix provides a full description of these methods, the simulation study, and provides links to the source code (www.stat.lsa.umich.edu/∼laber/UAI2008.html).

The confidence sets were constructed for sample sizes of $n = 30$ and $n = 50$ using the datasets given in table 1. All the data sets have binary labels and continuous features. The real datasets can be found at the UCI data repository (www.ics.uci.edu/mlearn). The simulated data sets were designed to investigate the performance of the new procedure in several scenarios of interest.

| Dataset | Motivation |
| --- | --- |
| Spam | Not Simulated |
| Ionosphere | Not Simulated |
| Heart | Not Simulated |
| Diabetes | Not Simulated |
| Abalone | Not Simulated |
| Liver | Not Simulated |
| Mammogram | Not Simulated |
| Magic | Not Simulated |
| Donut (Simulated) | $f^*$ close to Bayes |
| Outlier (Simulated) | $f^*$ far from Bayes |
| $\chi^2_{Small}$ (Simulated) | $f^* = 0$, low noise, far from Bayes |
| Three Points (Simulated) | $\hat{\xi}(\hat{f})$ highly non-normal, $f^*$ far from Bayes |

Table 1: Datasets used in experiments along with the reason for their inclusion. Here $f^*$ is the limiting value of the classifier as the amount of training data becomes infinite.

---

[2]Update: In the case of linear classification with squared error loss, current work in progress has proved this runs in polynomial time, on the order of $\mathcal{O}(n^{VC(\mathcal{G})})$.

[3]In this approach we are plugging in a resampled estimate and the effective sample size. That is, we are acting as if we had a test set.

For this series of experiments we fit a linear classifier using squared error loss. That is,

$$\mathcal{G} = \{f(x) = sign\big(\sum_{i=1}^{p} \beta_i \phi_i(x)\big), \beta_i \in \mathbb{R}\}$$

where the $\phi_i$ are transformations of the features (in all cases $\phi_i$ was either a polynomial or projection onto a number of principle components). The surrogate loss function is given by

$$\mathcal{L}(\mathcal{D}, f(;\boldsymbol{\beta})) = \sum_{(x_i, y_i) \in \mathcal{D}} \big(\sum_{j=1}^{p} \beta_j \phi_j(x_i) - y_i\big)^2,$$

and the scaling factor $\alpha_n$ used in $g$ is chosen to be $n$. For each data set the following procedure was used to estimate coverage probabilities.

1. Randomly divide data into training set $\mathcal{D}$ and testing set $\mathcal{T}$.
   - If the data set is real, randomly select $n$ training points for $\mathcal{D}$ and use the remainder for $\mathcal{T}$.
   - If the data set is simulated, generate $n$ data points for $\mathcal{D}$ and generate 5000 data points for $\mathcal{T}$.

2. Build classifier $\hat{f} = \arg\min_{f \in \mathcal{G}} \mathcal{L}(\mathcal{D}, f)$ using training data $\mathcal{D}$.

3. For each procedure use the training data $\mathcal{D}$ and chosen classifier $\hat{f}$ to construct confidence set with target coverage .950. The confidence interval is centered at $\hat{\xi}_\mathcal{D}(\hat{f})$ taken to be the .632 estimate [Efron 1983] in the the competing methods and the training error in the CUD bound.[4]

4. Using the $\xi(\hat{f}) = \hat{\xi}_\mathcal{T}(\hat{f})$ we record the coverage for each procedure.

The above steps are repeated 1000 times for each data set and the average coverage computed. The results are shown in tables 2 and 3 with the following abbreviations: quantile bootstrap (BS1) [Efron 1994], corrected boostrap (BS2) [Efron 1994], Yang's CV estimate (Y) [Yang 2006], Kohavi's normal approximation (K) [Kohavi 1996], a generalization of a Bayesian approach from Martin (M) [Martin 1996], and the inverted binomial of Langford (L) [Langford 2005].

The results for the $n = 50$ case are similar and the results are given in tables 4 and 5 of the appendix.

The results in tables 2 and 3 show promising results for the new method. It was the only method to provide the desired coverage on all ten datasets and, in addition, possessed the smallest diameter for two of them. The CUD bound confidence set is conservative, as must be expected by its construction. However, the diameter of the new confidence set is far from trivial, even for this extremely small sample size.

---

[4]The competing methods experience a significant reduction in performance if they are centered at the training error.

|  | n = 30 | | | | | | |
|---|---|---|---|---|---|---|---|
| Data | CUD | BS1 | BS2 | K | M | Y | L |
| Spam | **1.00** | .478 | **.983** | .782 | .632 | **.996** | .636 |
| Donut | **.999** | .880 | .908 | .631 | .633 | .937 | .620 |
| Ionosphere | **.998** | .605 | .926 | .816 | .757 | **.994** | .747 |
| Heart | **.999** | .406 | **.981** | .718 | .475 | **.998** | .458 |
| Diabetes | **1.00** | .654 | .900 | .912 | **.986** | **.997** | **.984** |
| Outlier | **.988** | .884 | .736 | .808 | .838 | .910 | **.961** |
| 3 Pt. | **.993** | .827 | .717 | .844 | .896 | .753 | **.952** |
| Abalone | **.983** | **.963** | .737 | **.972** | **.991** | **.998** | **.997** |
| Liv. | **.961** | **.964** | .878 | **.980** | **1.00** | **1.00** | **1.00** |
| Magic | **.998** | .918 | .920 | **.961** | **.982** | **.991** | **.992** |
| $\chi^2$ | **.985** | **.958** | .786 | **.961** | **.982** | **.996** | **1.00** |
| Mammogram | **1.00** | .678 | **.994** | .646 | .426 | **.983** | .406 |

Table 2: Average coverage of new method and competitors. Target coverage level was .950, those meeting or exceeding this level are highlighted in orange.

|  | n = 30 | | | | | | |
|---|---|---|---|---|---|---|---|
| Data | CUD | BS1 | BS2 | K | M | Y | L |
| Spam | .469 | .432 | **.432** | .315 | .314 | .451 | .354 |
| Donut | **.485** | .590 | .590 | .324 | .324 | .414 | .364 |
| Ionosphere | **.437** | .429 | .429 | .301 | .296 | .501 | .337 |
| Heart | .459 | .468 | .468 | .319 | .319 | **.433** | .359 |
| Diabetes | .433 | .305 | .305 | .307 | **.310** | .312 | .350 |
| Outlier | .555 | .491 | .491 | .328 | .329 | .456 | **.368** |
| 3 Pt. | .508 | .476 | .476 | .318 | .317 | .457 | **.356** |
| Abalone | .617 | **.314** | .314 | .331 | .331 | .504 | .371 |
| Liver | .559 | .372 | .372 | .327 | **.326** | .485 | .366 |
| Magic | .606 | .307 | .307 | .300 | **.283** | .464 | .324 |
| $\chi^2$ | .595 | **.330** | .330 | .329 | .330 | .501 | .351 |
| Mammogram | .464 | .532 | .532 | .321 | .321 | **.421** | .361 |

Table 3: Average diameter of confidence set constructed using new method and competitors. Method with smallest diameter and having correct coverage is highlighted in yellow.

It is also of interest to learn when competing methods perform poorly. Figure 1 plots the empirical coverage of the top four methods against the mean absolute value of the training error's kurtosis (standardized central fourth moment subtract 3) for each of the nine data sets listed in table 1. The kurtosis for a normal distribution is 0, and we use absolute kurtosis as a measure of deviation from normality. Figure 1 shows a trend of decreasing coverage as deviation from normality increases for the three distribution based methods. This suggests these methods may not be robust to non-normal training error. In particular we see that the method K (which has strongest normality assumptions) is most sensitive to deviation from normality. In contrast, the CUD bound method has stable, if conservative, coverage, regardless of the normality of training error.

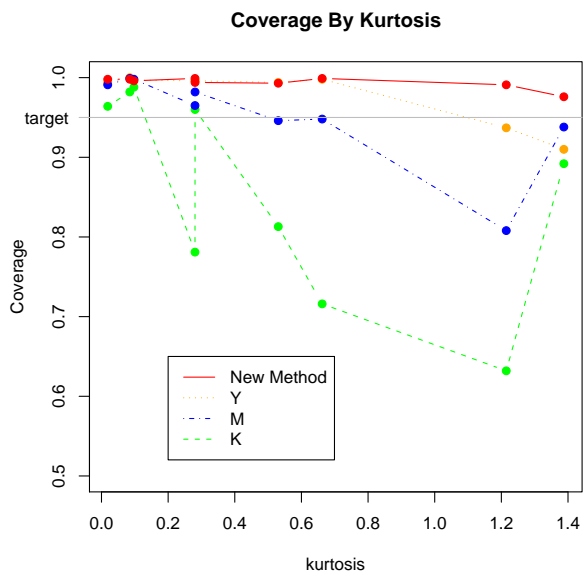

Figure 1: Shows coverage of top four methods by absolute value of kurtosis of the training error.

## 6  Conclusions and Future Work

In this paper we introduced a new method for constructing confidence sets for the generalization error in the small sample setting in classification. This bound is derived under the minimal assumption of a training set of *iid* draws from some unknown distribution $\mathcal{F}$. In addition, we provided a computationally efficient algorithm for constructing this new confidence set when the classifier is approximated by a parametric additive model. In preliminary experiments, the new confidence set exhibited superior coverage while maintaining a small diameter on a catalog of test and simulated data sets. Moreover, it was demonstrated that confidence sets based on distributional assumptions may not be robust to deviation from normality in the training samples.

Much work remains to be done. First, we need to provide theoretical guarantees for the level of confidence. We conjecture that a correct choice of $\alpha_n$ will depend on the complexity of the approximation space used to construct the classifier. It is well known that bootstrap estimators perform better when estimating smooth functions. Selecting $\alpha_n = \infty$ is similar to standard quantile bootstrap (BS1) which is ineffective in the classification setting because it bootstraps the non-smooth 0-1 loss. This suggests an upper bound on $\alpha_n$. Conversely, a smaller choice of $\alpha_n$ leads to a bootstrapped estimate of a smoother quantity but also introduces conservatism. For example, setting $\alpha_n = 0$ is equivalent to a uniform deviation bound on the training error which can often be trivially loose. This suggests a lower bound on $\alpha_n$ as well. Thus, $\alpha_n$ must strike a balance between smoothness required by the bootstrap and conservatism.

# Appendix: Experimental Results for $n = 50$

|  | n = 50 | | | | | | |
| --- | --- | --- | --- | --- | --- | --- | --- |
| Data | CUD | BS1 | BS2 | K | M | Y | L |
| Spam | 1.00 | .172 | .987 | .415 | .632 | .996 | .636 |
| Donut | .998 | .837 | .904 | .450 | .633 | .882 | .620 |
| Ionosphere | 1.00 | .292 | .975 | .575 | .757 | .985 | .747 |
| Heart | .999 | .072 | .999 | .186 | .475 | .998 | .458 |
| Diabetes | .999 | .654 | .900 | .912 | .986 | .984 | .984 |
| Out. | .984 | .805 | .893 | .702 | .838 | .933 | .961 |
| 3 Pt. | .972 | .800 | .812 | .648 | .896 | .624 | .952 |
| Abalone | .988 | .960 | .832 | .962 | .991 | .999 | .997 |
| Liver | .986 | .888 | .888 | .974 | 1.00 | .988 | 1.00 |
| Magic | 1.00 | .824 | .823 | .950 | .982 | .998 | .992 |
| $\chi^2$ | .999 | .955 | .791 | .960 | .982 | .996 | 1.00 |
| Mammogram | .999 | .317 | .999 | .146 | .426 | .991 | .406 |

Table 4: Average coverage of new method and competitors. Target coverage level was .950, those meeting or exceeding this level are highlighted in orange.

|  | n = 50 | | | | | | |
| --- | --- | --- | --- | --- | --- | --- | --- |
| Data | CUD | BS1 | BS2 | K | M | Y | L |
| Spam | .418 | .377 | .377 | .250 | .314 | .329 | .354 |
| Donut | .448 | .529 | .529 | .258 | .324 | .331 | .364 |
| Ionosphere | .386 | .364 | .364 | .237 | .296 | .315 | .337 |
| Heart | .409 | .436 | .436 | .254 | .319 | .443 | .359 |
| Diabetes | .433 | .365 | .365 | .247 | .310 | .358 | .350 |
| Out. | .442 | .439 | .439 | .262 | .329 | .328 | .368 |
| 3 Pt. | .440 | .413 | .413 | .253 | .317 | .357 | .356 |
| Abalone | .523 | .248 | .248 | .264 | .331 | .391 | .371 |
| Liver | .499 | .304 | .304 | .260 | .326 | .382 | .366 |
| Magic | .561 | .209 | .209 | .227 | .283 | .350 | .324 |
| $\chi^2$ | .401 | .265 | .265 | .236 | .330 | .389 | .351 |
| Mammogram | .419 | .511 | .511 | .255 | .321 | .320 | .361 |

Table 5: Average diameter of confidence set constructed using new method and competitors. Method with smallest diameter and having correct coverage is highlighted in yellow.